\documentclass{article}

\usepackage{amsmath, amsfonts, amssymb, amsthm}
\usepackage{txfonts} 
\usepackage{verbatim}
\usepackage{url}
\usepackage{color}
\usepackage{algorithm2e}
\usepackage[noend]{algpseudocode}
\usepackage{multirow}

\usepackage{siunitx}
\ifx\BlackBox\undefined
\newcommand{\BlackBox}{\rule{1.5ex}{1.5ex}}  
\fi
\ifx\proof\undefined
\newenvironment{proof}{\par\noindent{\bf Proof\ }}{\hfill\BlackBox\\[2mm]}
\fi

\newcommand\shortsection[1]{\vspace{6pt}{\noindent\bf #1.}}

\theoremstyle{definition}

\makeatletter
\def\url@leostyle{%
  \@ifundefined{selectfont}{\def\UrlFont{\sf}}{\def\UrlFont{\small\sffamily}}}
\makeatother
\makeatletter
\def\url@beostyle{%
  \@ifundefined{selectfont}{\def\UrlFont{\sf}}{\def\UrlFont{\scriptsize\sffamily}}}
\makeatother

\newcommand{\redacted}[1]{{\emph{$<$Redacted for Anonymity$>$}}}

\newcommand{\boneage}{RSNA Bone Age}
\newcommand{\graph}{ogbn-arxiv}
\newcommand{\census}{Census}

\newcommand{\gzero}{$\mathcal{G}_0$}
\newcommand{\gone}{$\mathcal{G}_1$}

\usepackage{microtype}
\usepackage{graphicx}
\usepackage{booktabs} 

\usepackage{hyperref}
\hypersetup{
    allcolors=., 
    }
\usepackage{amsthm}
\usepackage{amsfonts}
\usepackage{caption}
\usepackage{subcaption}
\usepackage{nicefrac}       

\theoremstyle{definition}
\usepackage{dashbox}
\usepackage[
    n,
    operators,
    advantage,
    sets,
    adversary,
    landau,
    probability,
    notions,    
    logic,
    ff,
    mm,
    primitives,
    events,
    complexity,
    asymptotics,
    keys]{cryptocode}
\usepackage{amssymb}
\usepackage{caption}
\usepackage{cleveref}

\usepackage{hyperref}

\usepackage[final, nonatbib]{neurips2021}
\usepackage[numbers]{natbib}

\newcommand{\papertitle}{Formalizing Distribution Inference Risks}

\title{\papertitle}

\author{Anshuman Suri and David Evans \\
Department of Computer Science\\
University of Virginia\\
\textsf{\small \{anshuman, evans\}@virginia.edu}}

\begin{document}

\maketitle

\begin{abstract}
Property inference attacks reveal statistical properties about a training set but are difficult to distinguish from the primary purposes of statistical machine learning, which is to produce models that capture statistical properties about a distribution. 
Motivated by Yeom et al.'s membership inference framework, we propose a formal and generic definition of property inference attacks. The proposed notion describes attacks that can distinguish between possible training distributions, extending beyond previous property inference attacks that infer the ratio of a particular type of data in the training data set. 
In this paper, we show how our definition captures previous property inference attacks as well as a new attack that reveals the average degree of nodes of a training graph and report on experiments giving insight into the potential risks of property inference attacks.
\end{abstract}

\section{Introduction}
\label{sec:introduction}

Inference attacks seek to infer sensitive information about the training process of a revealed machine-learned model, most often about the training data. For example, in a membership inference attack~\citep{shokri2017membership}, the adversary aims to infer whether a particular datum was part of the training data. 
In a property inference attack, an adversary aims to infer some statistical property of the training dataset, such as the proportion of women in a smile-detection dataset~\citep{ateniese2015hacking}. 

The research community lacks a good understanding of property inference attacks---for example, different notions are used in \citet{gopinath2019property} and \citet{zhang2020dataset}. Many formal privacy notions have been proposed, including numerous variations on differential privacy, but most notions focus on protecting specific data elements, not statistical properties of a dataset. One notable exception is the Pufferfish framework~\citep{kifer2014pufferfish}, which introduces notions that allow capturing aggregates of records via specifications of potential secrets and their relations. Although Pufferfish supports aggregate secrets, it is unclear how it can extend to distribution-level properties.
A recent attempt to formalize property inference~\citep{saeed} consists of a framework that reduces property inference to Boolean functions of individual members, posing the ratio of members satisfying the given function in a dataset as the property. However, such a formulation limits the threat model since it cannot capture many other kinds of statistical properties of the training distribution that may be sensitive, like the degree distribution of a graph~\citep{hay2009accurate}.

In this work\footnote{An expanded version of this work is also available~\cite{suri2021formalizing}}, we formalize property inference attacks based on the critical insight that the key difference between property inference attacks and other inference attacks is that the adversary's goal in the former is to learn about the training distribution, not the specific training dataset. Dataset inference attacks, such as membership inference~\citep{shokri2017membership} and attribute inference~\citep{fredrikson2014privacy}, operate on the level of training records. They are directly connected to the definition of differential privacy which bounds the ability to distinguish neighboring datasets. By contrast, distribution inference attacks attempt to learn statistical properties of the underlying distribution from which the training dataset is sampled.

\shortsection{Contributions}
We propose a simple and general experiment to formalize property inference attacks, inspired by \citet{yeom2018privacy}'s membership inference definition (Section~\ref{sec:property_defns}). Our definition is generic enough to capture any property of the underlying distribution, including, but not limited to, the attribute ratios that are the focus of previous property inference attacks. Motivated by this definition, we describe some experiments (Section~\ref{sec:experiments}) showing a variety of properties that fit our definitions and can be inferred by property inference attacks.

\section{Formalizing Property Inference}
\label{sec:property_defns}

Let $\mathcal{D}=(\mathcal{X}, \mathcal{Y})$ be a public distribution between data $\mathcal{X}$ and its corresponding labels $\mathcal{Y}$. We assume both the model trainer $\mathcal{B}$ and adversary $\mathcal{A}$ have access to $\mathcal{D}$. Both parties also have access to two functions that transform distributions: \gzero,\ \gone. Inferring properties of the distribution can reveal sensitive information in many scenarios, which can be captured by suitable choices of \gzero\ and \gone.

In our cryptographic game definition, $\mathcal{B}$ picks one of these distribution transformers at random and samples a dataset $D$ from the resulting distribution.
Given access to a model $M$ trained on $D$, the adversary aims to infer which of the two distribution mappers is used. The property inference experiment can be described as: 
\begin{figure}[h]
    \centering
    \fbox{%
        \pseudocode[]{%
            \textbf{Trainer } \bdv \<\< \textbf{Adversary } \adv \\[][\hline]
             \\
            \pcln b \sample \bin \\
            \pcln D \sim \mathcal{G}_b(\mathcal{D}) \\
            \pcln M \xleftarrow{train}{} D  \\
            \pcln \< \sendmessageright{top={$M$}, length=2cm} \\
            \pcln \< \< \hat{b} = \mathcal{H}(M)
        }
    }
\end{figure}

If $\adv$ can successfully infer $b$ via $\hat{b}$, then it can determine which of the two properties the training distribution satisfied. The advantage of the adversary $\adv$ using algorithm $\mathcal{H}$ is:
$$    \text{Adv}_{\mathcal{H}} = \abs{\condprob{\hat{b}}{b} - \condprob{\hat{b}}{\neg\ b}}.
$$
This advantage is negligible when the adversary does no better than random guessing.

\shortsection{Applying the Definition}
\label{sec:existing_defns}
Seminal works on property inference~\citep{ateniese2015hacking,ganju2018property} involve a model trained either on the original dataset or a version modified to bias towards some chosen attribute. Our definition can be used to describe these attacks by setting \gzero\ to the identity function (so $\mathcal{G}_0(\mathcal{D})$ is original distribution $\mathcal{D}$) and \gone\ to a filter that adjusts the distribution to have a specified ratio over the desired attribute.
With respect to a binary property function $f:\mathcal{X} \xrightarrow{} \bin$, $\mathcal{D}$ can be characterized using a generative probability density function:
\begin{align}
    \rho_{\mathcal{D}}(\textbf{x}) = \sum_{c \in \bin}p(c)\cdot p(\textbf{x}\;|\;c),
\end{align}
where $p(c)$ is a multinomial distribution representing the probabilities over the desired (binary) property function $f$ and its possible values $c$, and $p(\textbf{x}|c)$ is the generative conditional probability density function. Then, \gone$(\mathcal{D})$ can be expressed using the following probability density function (with an adjusted prior $\hat{p}$):
\begin{align}
    \rho_{G_1(\mathcal{D})}(\textbf{x}) = \sum_{c \in \bin}\hat{p}(c)\cdot p(\textbf{x}\;|\;c) &\ ,\ \hat{p}(1) = \alpha \label{eq:alpha}
\end{align}
where $\alpha$ is the probability of a randomly sampled point satisfying the property function $f$. Thus, a uniformly randomly sampled dataset from \gone$(\mathcal{D})$ would have an expected ratio $\alpha$ of its members satisfy $f$. Additionally, we can modify \gzero\ with a similarly adjusted prior, enabling the adversary to distinguish between two chosen ratios~\citep{zhang2020dataset}.

\section{Experiments}
\label{sec:experiments}

To better understand how well an intuitive notion of divergence in properties aligns with observed inference risk, we execute property inference attacks with increasing diverging properties on tabular, image, and graph datasets. 
Code for our experiments is available at \url{https://github.com/iamgroot42/distribution_inference}.

\subsection{Datasets}
\label{sec:datasets}

We report on experiments using three datasets. For each dataset, we construct non-overlapping data splits between $\adv$ and $\bdv$. Both parties then modify their data according to the desired property to emulate a distribution: sampling datasets from these partitions to train and evaluate their models.

\textbf{\census}~\citep{bay2000uci} consists of several categorical and numerical attributes like age, race, education level to predict whether a person makes over \$50K a year. We focus on the ratios of whites (race) and females (sex) as properties.

\textbf{\boneage}~\citep{halabi2019rsna} contains X-Ray images of hands, with the task being predicting the patients' age in months. We convert the task to binary classification based on an age threshold and use a pre-trained DenseNet~\citep{huang2017densely} model for feature extraction. We focus on the ratios of the females (available as metadata) as properties.

\textbf{\graph}~\citep{wang2020microsoft} is a directed graph, representing citations between computer science arXiv papers. The task is to predict the subject area categories for unlabeled papers. We infer the mean node-degree property of the graph.

Since the original ratios may be unbalanced (Section~\ref{sec:existing_defns}), we fix \gzero\ such that the resulting distribution is balanced (0.5 ratio) for the chosen attribute for the Census and \boneage~datasets., setting \gone\ based on varying $\alpha$ values substituted in Equation~\ref{eq:alpha}. For the \graph~dataset, we set \gzero\  such that the graph has a mean node-degree of 13. For \gone, we modify the graph to have a mean-degree $\alpha$.

\subsection{Attacks}
\label{sec:attacks}

We use two simple attacks and the state-of-the-art meta-classifier attack in our experiments.

\shortsection{Loss Test}
A simple algorithm $\mathcal{H}$ consists of using the loss function $l$ used for training the model $M$ to compute $\hat{b}$. For datasets $D_0 \sim \mathcal{G}_0(\mathcal{D})$, $D_1 \sim \mathcal{G}_1(\mathcal{D})$, 
$\hat{b} = \mathbb{I}[l(M, D_0) > l(M, D_1)]$ 
where $\mathbb{I}$ is the indicator function. Intuitively, a model would have lower test loss on data sampled from the training distribution, compared to another distribution. This method does not require the adversary to train models, but only to have access to suitable test distributions.

\shortsection{Threshold Test}
The loss test assumption may not hold for some pairs of properties if one distribution is inherently easier to classify than the other (as we observe in our experiments in Section~\ref{sec:results}).
The adversary can train models on datasets sampled from each distribution and observe their performance on $D_0$, $D_1$ to 
derive a threshold $\lambda$ to maximize accuracy distinguishing between the two($D_k$ yields higher accuracy). $\hat{b}$ is then computed as : $\hat{b} = \mathbb{I}[l(M, D_k) > \lambda]$.

\shortsection{Meta-Classifiers}
The state-of-the-art property inference attack is \citet{ganju2018property}'s attack using permutation-invariant networks as meta-classifiers. The meta-classifiers take as input model parameters (weights, bias) and predict which of the distributions was used to train the model. 
Since the meta-classifier is itself a classifier that requires many (we use 1000 for each distribution in our experiments) models trained on the two distributions to train, this attack is only feasible for adversaries with access to both training distributions and considerable computational resources.

\subsection{Results}
\label{sec:results}

\begin{figure}[b]
\centering
\includegraphics[width=0.33\textwidth]{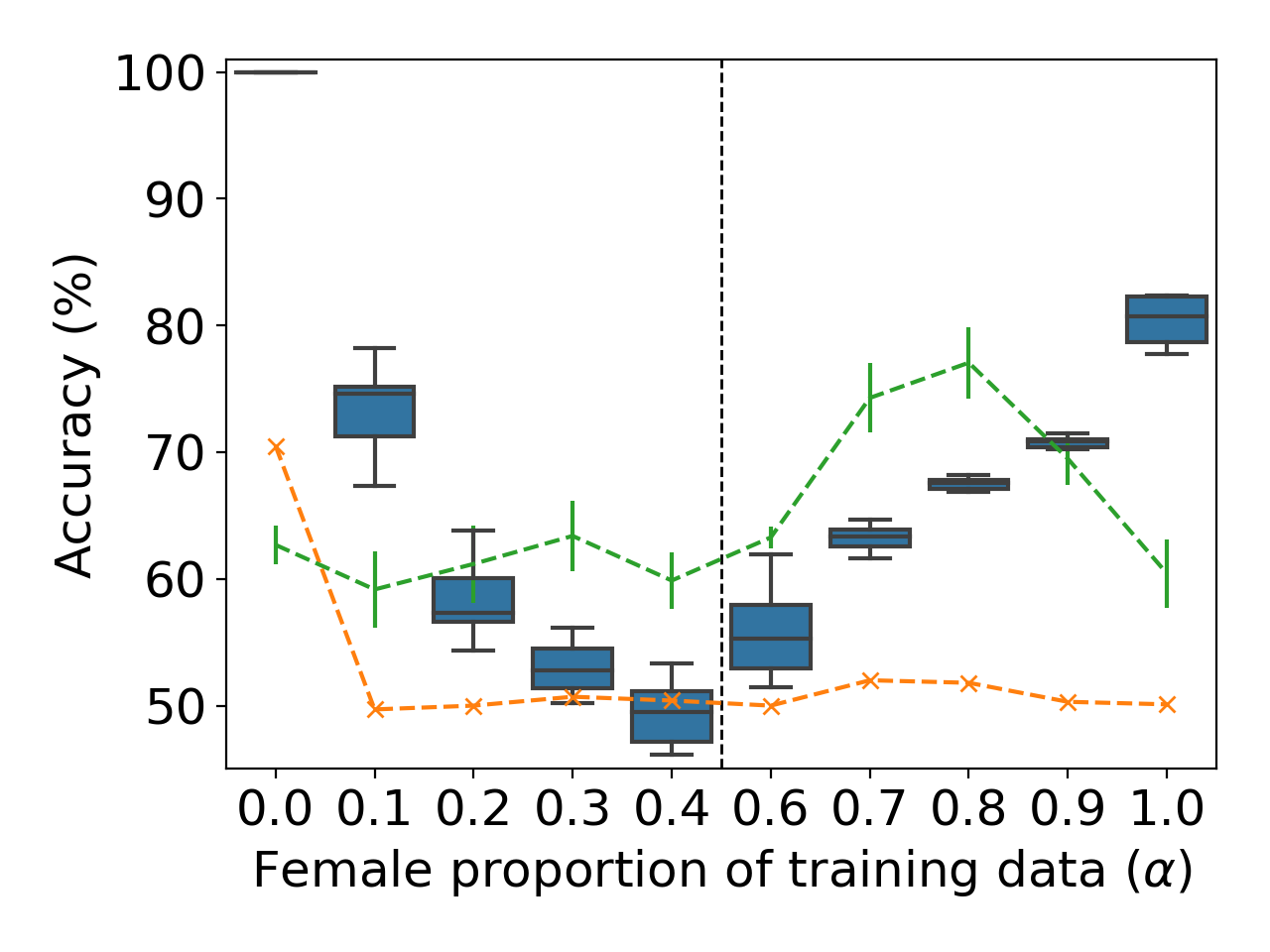}
\includegraphics[width=0.33\textwidth]{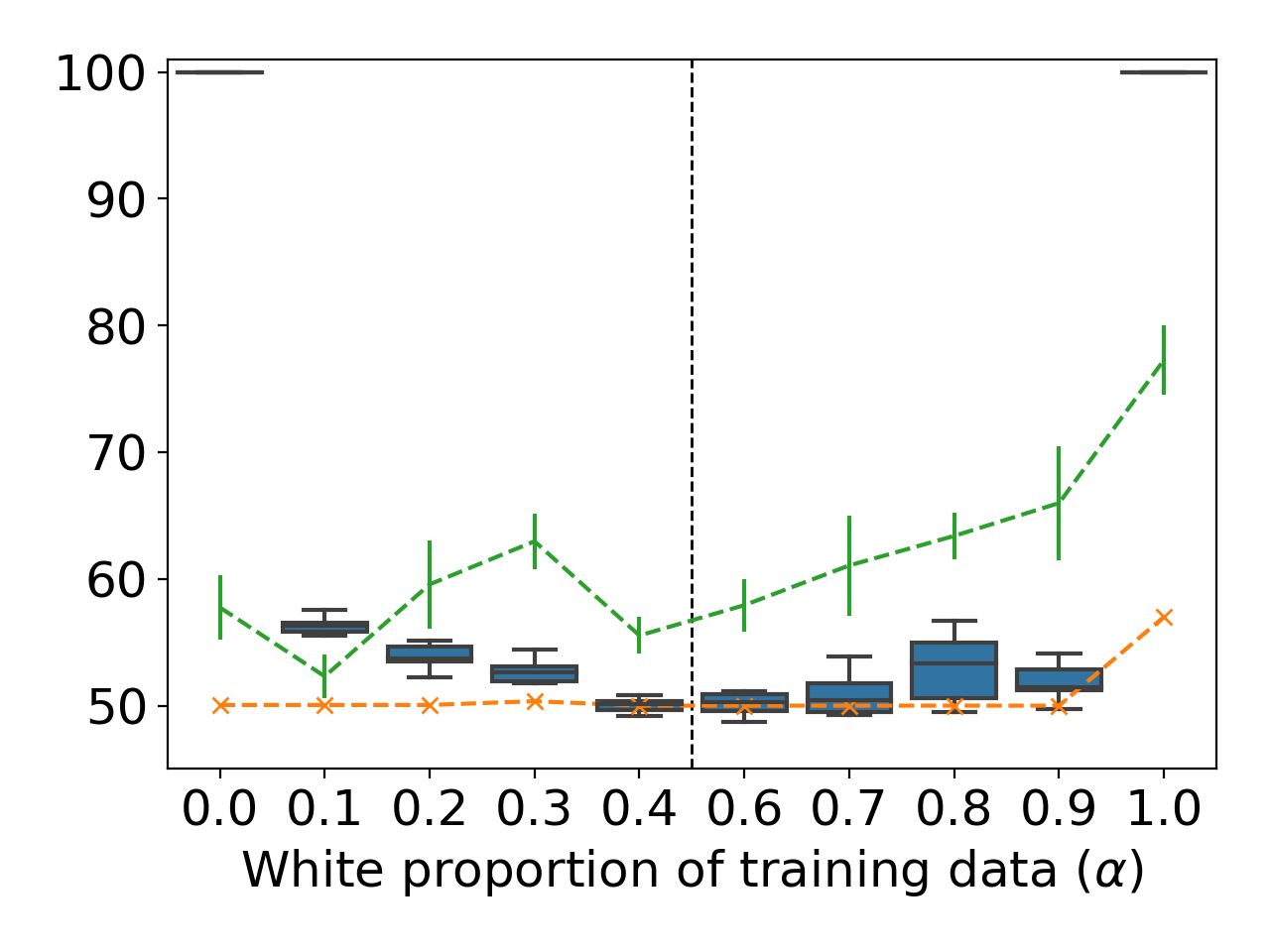}
\includegraphics[width=0.33\textwidth]{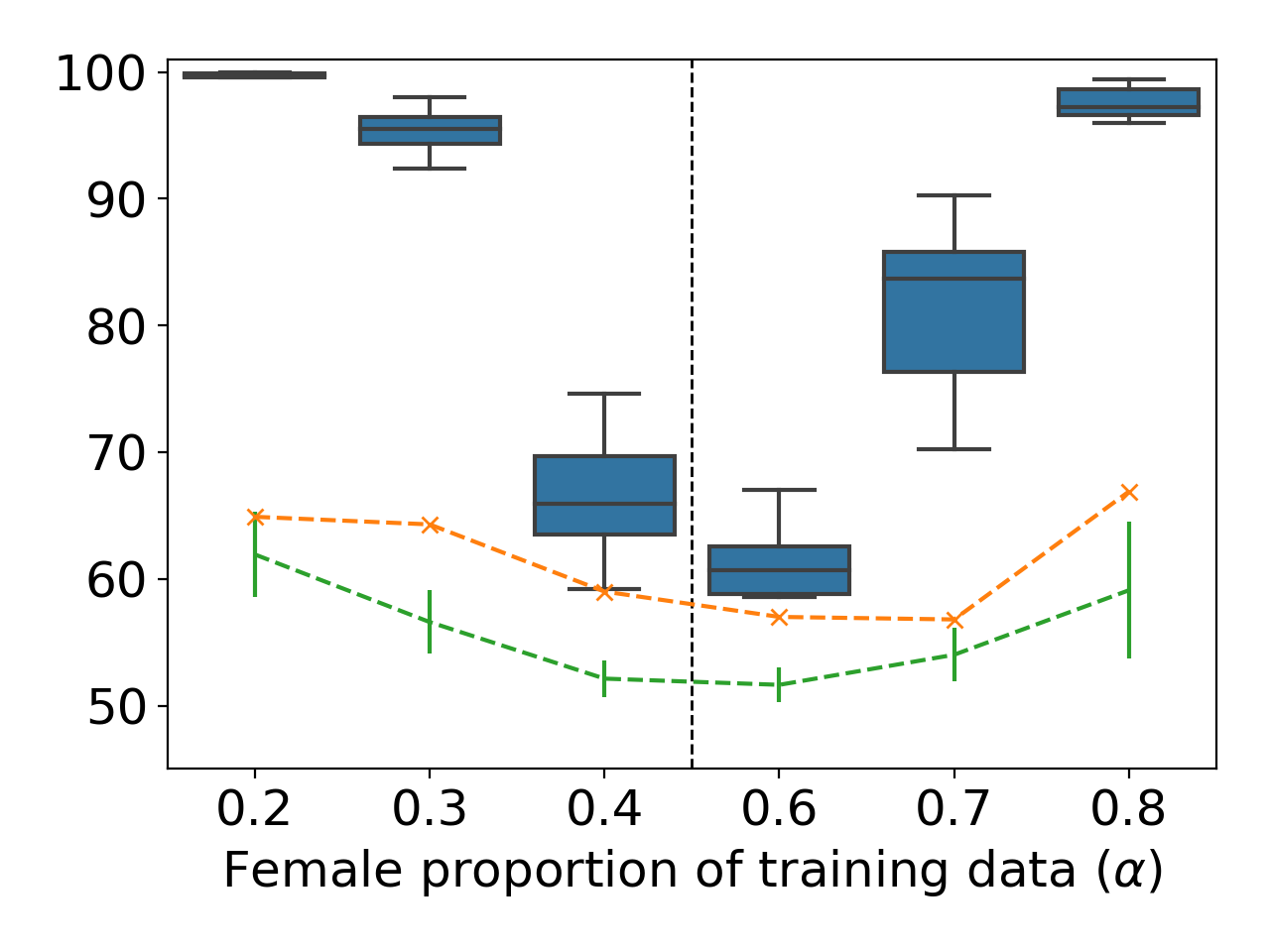}

\caption{Classification accuracy for distinguishing between training distributions for unseen models for on the \census\ (females: left, whites: middle) and \boneage\ (right) datasets. Orange crosses are for the Loss Test; green with error bars are the  Threshold Test; the blue box-plots are the meta-classifiers.}
\label{fig:census_and_boneage}
\end{figure}

\begin{figure}[tb]
\centering
\includegraphics[width=0.36\textwidth]{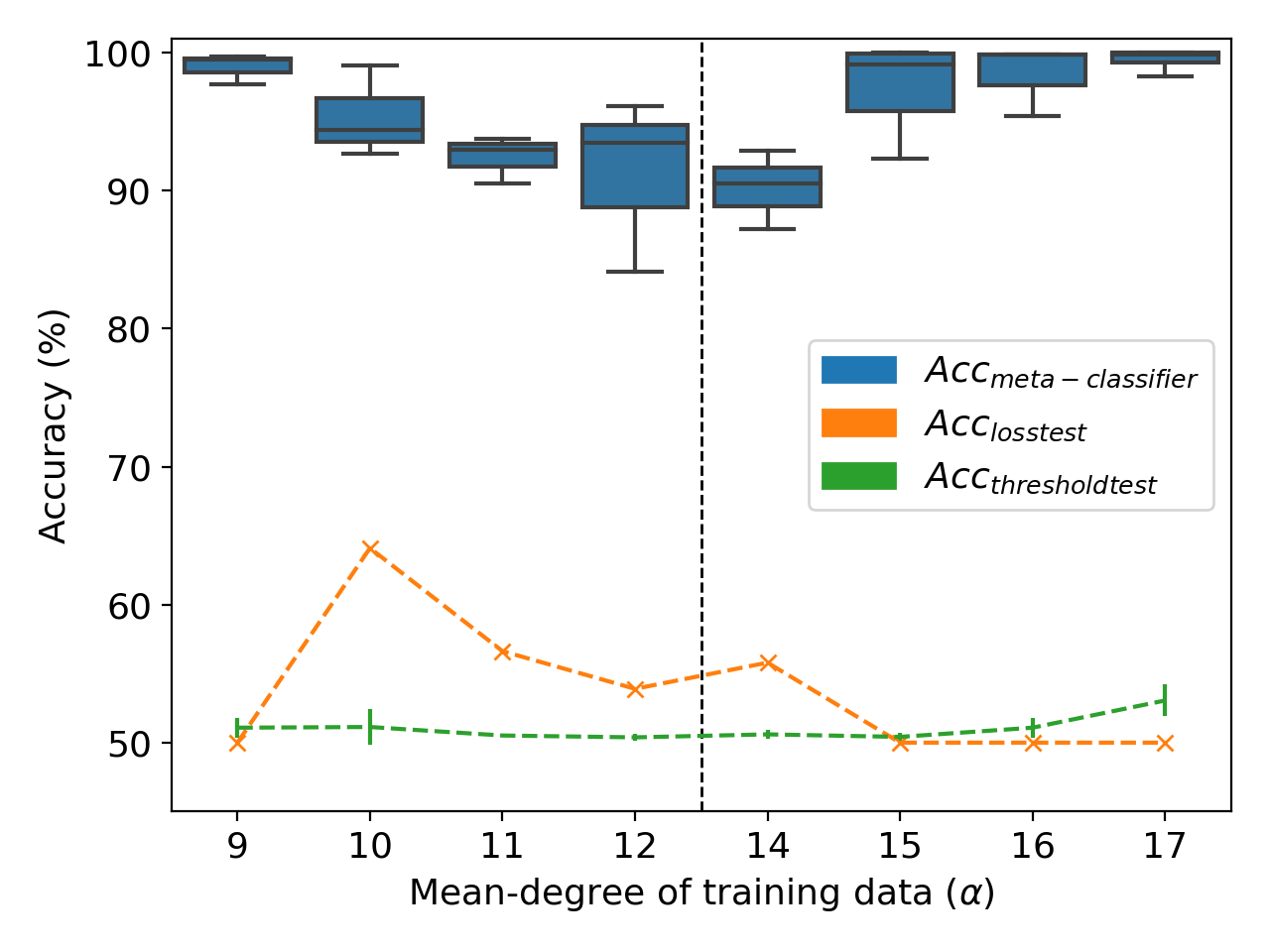}
\includegraphics[width=0.40\textwidth]{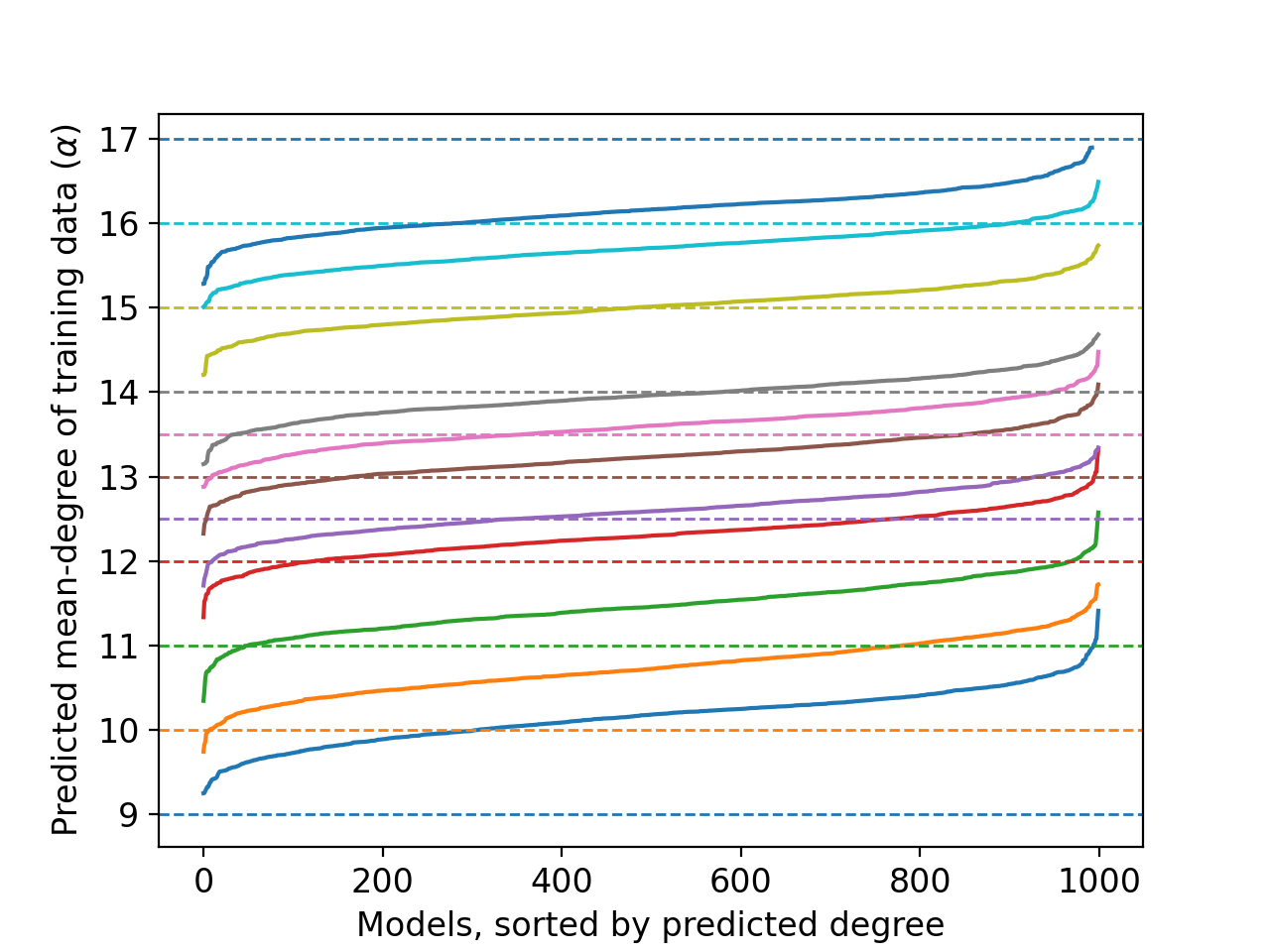}
\caption{Performance on unseen models on the \graph\ dataset, for binary classification (left) and direct $\alpha$ prediction via regression (right). For regression, each color represents the true degree (dashed lines) of the models  being tested.}
\label{fig:boneage_and_arxiv}
\end{figure}

We try the attacks on a range of diverging distributions to evaluate inference risks and understand how well an adversary could distinguish models trained using different distributions. Figures~\ref{fig:census_and_boneage} and \ref{fig:boneage_and_arxiv} summarize results on the three datasets.

\shortsection{\census}
The loss test performs marginally better than random in most cases, while the threshold test often outperforms meta-classifiers (left two sub-figures of Figure~\ref{fig:census_and_boneage}). We observe a clear trend with meta-classifiers while varying $\alpha$ for the sex attribute: performance improves as distributions diverge ($\alpha$ further away from 0.5). However, this trend is not symmetric around \gzero$(\mathcal{D})$. For example, the adversary performs better on $\alpha=0.0$ than $\alpha=1.0$, even though both are equally divergent from \gzero$(\mathcal{D})$ with respect to female proportion. Even more surprising is the case for race. The meta-classifiers barely do better than random guessing except at the two extremes (0.0 (no whites) and 1.0 (all whites)).

\citet{ganju2018property} applied their meta-classifier method on two properties on this dataset: 38\% v/s 65\% women (case A), and 0\% v/s 87\% whites (case B). For case B, the loss-test performs as well as random guessing (50.1\%), while the threshold-test ($92.4\pm2.6$\%) approaches meta-classifier performance ($99.0\pm0.1$\%). For case A, the threshold-based method ($62.7 \pm 2.0$\%) out-performs the meta-classifiers ($62.1\pm 1.7$\%), while the loss test method fails (50\%).\footnote{Ganju \textit{et al.} report 97\% accuracy for case A and 100\% for case B. We were able to reproduce these results in the original setting, but conducted the reported experiments in a more realistic setting where we ensure there is no victim/adversary overlap and maintain the label ratios and same dataset sizes (Appendix~\ref{sec:appendix_attacks}). This leads to drops in meta-classifier accuracy, especially in case A, but reduces the risk that the inference is actually based on other differences between the distributions used, not just the inferred property.}

\shortsection{\boneage}
The meta-classifier method is able to distinguish the proportion of females in the training distribution, with a substantial adversarial advantage even when the proportions are close to 0.5 (rightmost subfigure in Figure~\ref{fig:census_and_boneage}). 
Accuracy improves as the distributions diverge, 
however, this trend is not symmetric---the meta-classifier is more accurate for $\alpha=0.3$ than $\alpha=0.7$, even though both are equally divergent from \gzero$(\mathcal{D})$.
The loss test performs marginally better than random guessing in some cases and yields non-trivial accuracies in others. For cases like $\alpha=0.4$, it even comes close to matching the meta-classifier performance, with a performance gap of less than $4\%$. However, this method's performance does not increase as distributions diverge from each other (Figure~\ref{fig:census_and_boneage}), and the threshold test consistently performs worse than the loss test.

\shortsection{\graph}
We vary $\alpha$ in [9, 17] at intervals of one, producing test datasets by pruning either high or low-degree nodes from the original graph to achieve a desired $\alpha$. Similar to \boneage, meta-classifier performance increases as the distributions diverge, albeit with much smaller drops. The meta-classifier is highly successful at differentiating between the two distributions: even when the difference in mean node-degrees is just one (Figure~\ref{fig:boneage_and_arxiv}). 
Both loss test and threshold test fail on this dataset.

We also trained a regression variant of the meta-classifier to predict $\alpha$ directly. We vary $\alpha$ in the same range. The resulting meta-classifier performs quite well, achieving a mean-squared error (MSE) loss of $0.393\pm0.360$ 
It generalizes well to unseen distributions, achieving an MSE loss of 0.076 on $\alpha=\{12.5, 13.5\}$. A property inference adversary can thus be strong enough not just to distinguish between two possible distributions, but to infer the actual average node degree of the training dataset. 

\section{Conclusions}

We propose a general definition for property inference attacks, focusing on distributions instead of datasets. Our definition supports arbitrary properties like the mean node-degree of a graph, which we demonstrate a successful adversary can infer. A principled approach to executing such attacks reveals how intuition may not necessarily align with actual performance---seemingly similar pairs of distributions can have starkly different attack success rates, and simple attacks can yield non-trivial accuracy. 
We expect there is room for improving attacks that do not rely on meta-classifiers, and studying the performance of such attacks will give more valuable insights into when properties can be inferred. 

\subsection*{Acknowledgements}
\noindent

This work was partially supported by grants from the National Science Foundation (\#1717950 and \#1915813).

\bibliography{main}
\bibliographystyle{abbrvnat}

\clearpage
\appendix
\section{Experimental Setup}
\label{sec:appendix_training_info}

\subsection{Data and Model Preparation}
\label{sec:appendix_data_prep}

To ensure a fair comparison between models, we ensure that sample sizes per dataset are comparable for all distributions considered in our attacks. Furthermore, these samples are randomly selected, ruling out the possibility of property inference attacks inadvertently performing some form of dataset inference: differentiating between exact datasets instead of their properties.

\shortsection{\census}
After converting categorical data to one-hot features, the dataset reduces to 42-feature vectors for each datapoint. While performing data splits (for adversary/victim and train/test), we use stratification with the sex/race attribute and class labels, thus ensuring that changing one of the ratios does not inadvertently change any other ratios. We train neural networks with three hidden layers of dimensions (32, 16, 8).

\shortsection{\boneage} We bin the age label into binary categories, based on whether the age is more than 132 months (label 1) or not (label 0). Picking this threshold leads to a balanced label distribution. Similar to \census, we perform stratified data splits, keeping the label ratios in mind. We train neural networks with two hidden layers of dimensions (128, 64)

\shortsection{\graph} We use the publication years to perform all node label splits. This year-based split ensures no overlap in the victim/adversary data or their train/test data. We pre-process the graph following the Open Graph Benchmark~\citep{NEURIPS2020_fb60d411}. Additionally, we add reverse edges in the graph to convert it to bidirectional, in addition to self-loops. To add randomness to the graph, we randomly prune 1\% of the nodes for each experiment. We trained Graph Convolutional Networks~\citep{kipf2016semi} with three convolutional layers of 256 channels each.

\subsection{Attack Details}
\label{sec:appendix_attacks}

We perform each experiment ten times and report mean values with standard deviation in all of our experiments. Since the loss test uses a fixed test set per experiment, there is no variation in its results. The model trainer produces 1000 models per distribution using its split of data. These models serve as the evaluation set. Thus, any particular experiment distinguishing between two distributions uses 2000 models for evaluation. The adversary trains 1000 models per distribution on its split of data for the threshold loss and meta-classifier experiments. 

\shortsection{Loss Test} The adversary uses its test data to sample the two test sets $D_0, D_1$. Since we use the same test data in evaluations, we turn off sampling (see above) for this setting.

\shortsection{Threshold Loss}
The adversary uses a small sample size of 50 models per distribution to identify which $D_k$ leads to a more significant performance gap between its models. After identifying $k$, it finds $\lambda$ that maximizes the classification accuracy between models trained on datasets from the two distributions, using a simple linear search. It then uses $k$ and $\lambda$ to generate its predictions.

\shortsection{Meta-Classifier} We used permutation invariant networks as our meta-classifier architecture~\citep{ganju2018property}. This architecture is invariant to different neuron orderings in a neural network and performs well even with a few thousand models. The adversary randomly samples 1600 models from its pool to train the meta-classifier. Following experimental designs from prior works, we were able to achieve the accuracies that the authors reported. However, we notice that, for some cases, using our experimental design leads to significant performance drops. 
Steps like ensuring no overlap in victim/adversary training data, randomly sampled datasets for \gzero$(\mathcal{D})$, \gone$(\mathcal{D})$, ensuring the same dataset size, introduce stochasticity that makes it for a meta-classifier to perform as well. Thus, these steps help emulate a more realistic scenario but adversely impact meta-classifier in some cases.

\end{document}